\documentclass[10pt,twocolumn,letterpaper]{article}

\usepackage[pagenumbers]{iccv} 

\definecolor{iccvblue}{rgb}{0.21,0.49,0.74}
\usepackage[pagebackref,breaklinks,colorlinks,allcolors=iccvblue]{hyperref}

\usepackage{multirow}
\usepackage[table,xcdraw]{xcolor}
\usepackage{adjustbox}
\usepackage[normalem]{ulem}
\useunder{\uline}{\ul}{}

\usepackage[accsupp]{axessibility}  

\def\confName{ICCV}
\def\confYear{2025}

\title{Fast Image Super-Resolution via Consistency Rectified Flow}

\author{
Jiaqi Xu$^{1,2} $\thanks{Part of work done as an intern at Huawei.} ,
Wenbo Li$^{2}$\thanks{Corresponding author.} ,
Haoze Sun$^{2}$,
Fan Li$^{2}$,
Zhixin Wang$^{2}$,
Long Peng$^{2}$,
Jingjing Ren$^{3}$, \\
Haoran Yang$^{1}$,
Xiaowei Hu$^{4}$, 
Renjing Pei$^{2}$\footnotemark[2] ,
and Pheng-Ann Heng$^{1}$
\vspace{1mm} \\
$^1$The Chinese University of Hong Kong
$^2$Huawei Noah's Ark Lab \\
$^3$HKUST (GZ)
$^4$South China University of Technology
}

\begin{document}
\maketitle
\begin{abstract}
Diffusion models (DMs) have demonstrated remarkable success in real-world image super-resolution (SR), yet their reliance on time-consuming multi-step sampling largely hinders their practical applications. While recent efforts have introduced few- or single-step solutions, existing methods either inefficiently model the process from noisy input or fail to fully exploit iterative generative priors, compromising the fidelity and quality of the reconstructed images. To address this issue, we propose FlowSR, a novel approach that reformulates the SR problem as a rectified flow from low-resolution (LR) to high-resolution (HR) images. Our method leverages an improved consistency learning strategy to enable high-quality SR in a single step. Specifically, we refine the original consistency distillation process by incorporating HR regularization, ensuring that the learned SR flow not only enforces self-consistency but also converges precisely to the ground-truth HR target. Furthermore, we introduce a fast-slow scheduling strategy, where adjacent timesteps for consistency learning are sampled from two distinct schedulers: a fast scheduler with fewer timesteps to improve efficiency, and a slow scheduler with more timesteps to capture fine-grained texture details.
Extensive experiments demonstrate that FlowSR achieves outstanding performance in both efficiency and image quality. Code: \href{https://github.com/jiaqixuac/FlowSR}{jiaqixuac/FlowSR}.

\end{abstract}
    
\section{Introduction}
\label{sec:flowsr_intro}

Real-world image super-resolution (SR) aims to reconstruct high-resolution (HR) images from their low-resolution (LR) counterparts while simultaneously removing unknown degradations. 
With the remarkable success of diffusion models (DMs) \cite{ho2020denoising,song2021score}, SR methods leveraging these models—particularly those built on powerful text-to-image (T2I) models like Stable Diffusion (SD) \cite{rombach2022high}—have demonstrated outstanding performance~\cite{wang2024exploiting,yu2024scaling,wu2024seesr}. 
However, these diffusion-based SR approaches require an iterative reverse sampling process that gradually refines noisy inputs into HR outputs. 
Despite advancements such as efficient ordinary differential equation (ODE) solvers like DDIM~\cite{song2021denoising}, the slow inference speed remains a bottleneck, limiting their practicality in real-world applications.

\begin{figure}[t]
    \centering
    \includegraphics[width=\hsize]{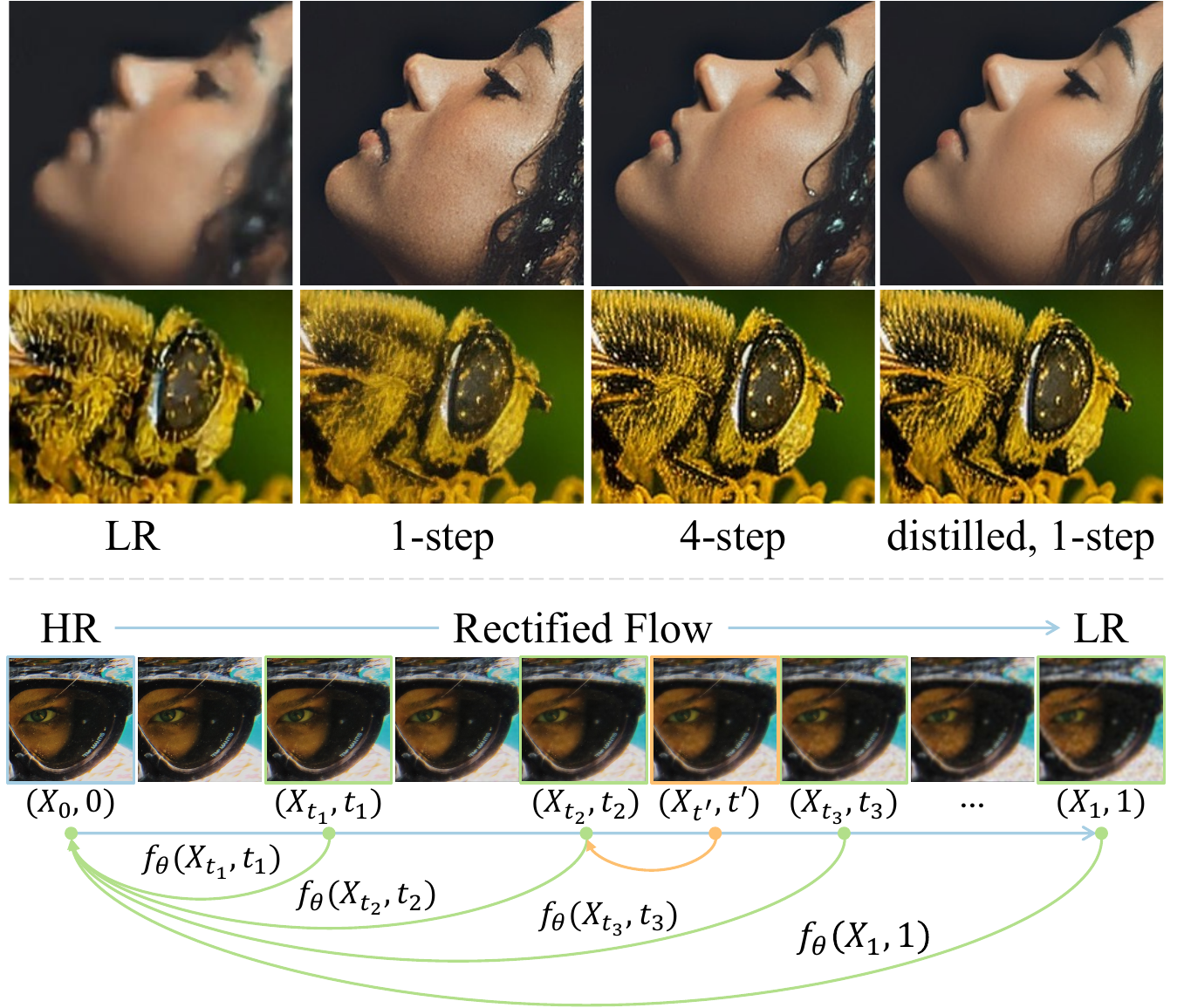}
    \caption
    {Our consistency SR flow model achieves high-quality single-step inference (distilled, 1-step) by distilling from multi-step higher-quality sampling process (top).
    We achieve this by formulating SR as a rectified flow that bridges LR and HR images, combined with improved consistency learning (bottom).}
    \label{fig:flowsr_intro}
    \vspace{-4mm}
\end{figure}

Recently, few-step or single-step SR methods derived from diffusion models are designed~\cite{yue2023resshift,cui2024taming,wang2024sinsr,wu2024one,zhang2024degradation}.
Several studies focus on designing more efficient diffusion processes for SR.
ResShift \cite{yue2023resshift} speeds up diffusion by shifting the LR–HR residual into the Markov chain and reduces sampling to 15 steps,
while DoSSR \cite{cui2024taming} introduces a diffusion process more compatible with pre-trained DMs.
However, they rely on DDPM~\cite{ho2020denoising}, which uses a more curved transition trajectory and starts with a noise-perturbed LR image that loses crucial information, ultimately limiting performance.
In parallel, another line of work targets single-step SR by learning the LR–HR mapping directly.
For example, SinSR \cite{wang2024sinsr} learns one-step SR prediction by leveraging the teacher's output from ResShift \cite{yue2023resshift} and ground-truth HR as target,
whereas OSEDiff adapts pre-trained SD for SR and refines one-step prediction with the VSD loss~\cite{wang2024prolificdreamer}.
However, these methods fail to fully harness the advantages of iterative generative modeling, which naturally facilitates high-quality texture synthesis.

In this work, we present FlowSR, which unifies rectified flow with consistency models (CMs) to enable efficient, single-step image super-resolution.
As illustrated in \cref{fig:flowsr_intro}, we reformulate SR as a rectified flow \cite{liu2022flow}, which establishes a simple and straight ODE-based mapping between LR and HR images.
By explicitly modeling this trajectory, FlowSR learns fine-grained LR-to-HR transformations for better SR quality and facilitates fast sampling.
Building upon this formulation, we further leverage consistency models \cite{song2023consistency} to enhance single-step inference.
Enforcing consistency across points on the same SR flow trajectory distills multi-step higher-quality restoration into fewer steps that reach to the same HR result.

However, the naive consistency distillation (CD) objective in CMs is suboptimal for SR flow, where the SR task demands both high quality and high fidelity.
While CD enforces self-consistency across discretization steps, there is no guarantee that the final distillation target aligns well with the ground-truth HR.
To address this, we propose HR-regularized consistency learning, which explicitly requires the model’s predictions to match real HR images.
This additional constraint mitigates teacher-induced errors in the distillation target and enhances the reconstruction of fine-grained details.
To further improve efficiency and robustness of the consistency SR flow model, we introduce a fast-slow time scheduling strategy.
Rather than sampling distillation timestep pairs from the two boundaries of a discretized interval, we sample adjacent timesteps from distinct ``fast'' and ``slow'' schedulers.
The fast scheduler uses fewer timesteps to facilitate efficient inference, while the slow scheduler employs more granular steps to maintain alignment with the SR flow objectives.
This mixed sampling introduces large and flexible jumps with mild perturbations into the HR regularization steps, allowing for a broader coverage of SR flow trajectories.

To train our consistency SR flow model, we first fine-tune a pre-trained SD model to align with the SR flow objective, followed by consistency SR flow distillation.
Additionally, we incorporate a GAN \cite{goodfellow2014generative} loss and an image quality alignment loss, with the latter promoting desirable text-described attributes in the restored images, thereby improving the SR quality.

As illustrated in \cref{fig:flowsr_intro}, our model leverages the proposed consistency SR flow distillation to transfer the quality improvement typically achieved with multiple sampling steps into a single step.
Experiments on real-world datasets demonstrate the superiority of FlowSR.
Our core contribution lie in a new paradigm to solve the one-step SR problem.
First, we explore efficient flow modeling for SR and identify potential challenges when incorporating consistency models.
Second, we introduce several techniques, including HR regularization in consistency learning and a fast-slow time scheduling to enhance one-step SR performance.

\section{Related Work}
\label{sec:flowsr_background}

\subsection{Image Super-Resolution}
Real-world image super-resolution (SR) aims to reconstruct high-resolution (HR) images from low-resolution (LR) inputs, tackling complex and unknown degradations like noise, blur, and compression artifacts.
Since \citet{dong2014learning}, numerous deep learning-based methods have been proposed to address the SR problem from different perspectives, including network design \cite{liang2021swinir}, model training \cite{ledig2017photo}, and degradation simulation~\cite{wang2021real,zhang2021designing,peng2025towards}.
Recently, diffusion models~\cite{ho2020denoising,song2021score} have achieved remarkable success in image generation~\cite{rombach2022high,ren2024ultrapixel}, which inspires their application to SR and demonstrate significant advancements~\cite{saharia2022image,wang2024exploiting}.

\vspace{-4mm}
\paragraph{Diffusion Model-Based SR}
Conditioned on the LR image, diffusion model-based SR iteratively denoises towards the target HR image.
The LR condition can be leveraged either by input concatenation~\cite{saharia2022image} or via adapters~\cite{zhang2023adding,mou2024t2i}.
StableSR~\cite{wang2024exploiting} based on Stable Diffusion~\cite{rombach2022high}, uses a trainable time-aware encoder to incorporates LR.
Subsequent improvements are achieved through enhanced LR conditioning in DiffBIR \cite{lin2024diffbir}, semantics-aware prompts in SeeSR \cite{wu2024seesr}, reference image generation in CoSeR \cite{sun2024coser}, and scaled-up training in SUPIR \cite{yu2024scaling}.
However, these DM-based SR approaches, starting from noise, suffer from low inference speed, typically requiring 50-200 sampling steps.
In contrast, ResShift~\cite{yue2023resshift} constructs an efficient diffusion model by shifting the residual between HR and LR. Denoising from the noised LR reduces the sampling steps to 15 but it still suffers from limited restoration quality.

\begin{figure*}[t]
    \centering
    \includegraphics[width=0.95\hsize]{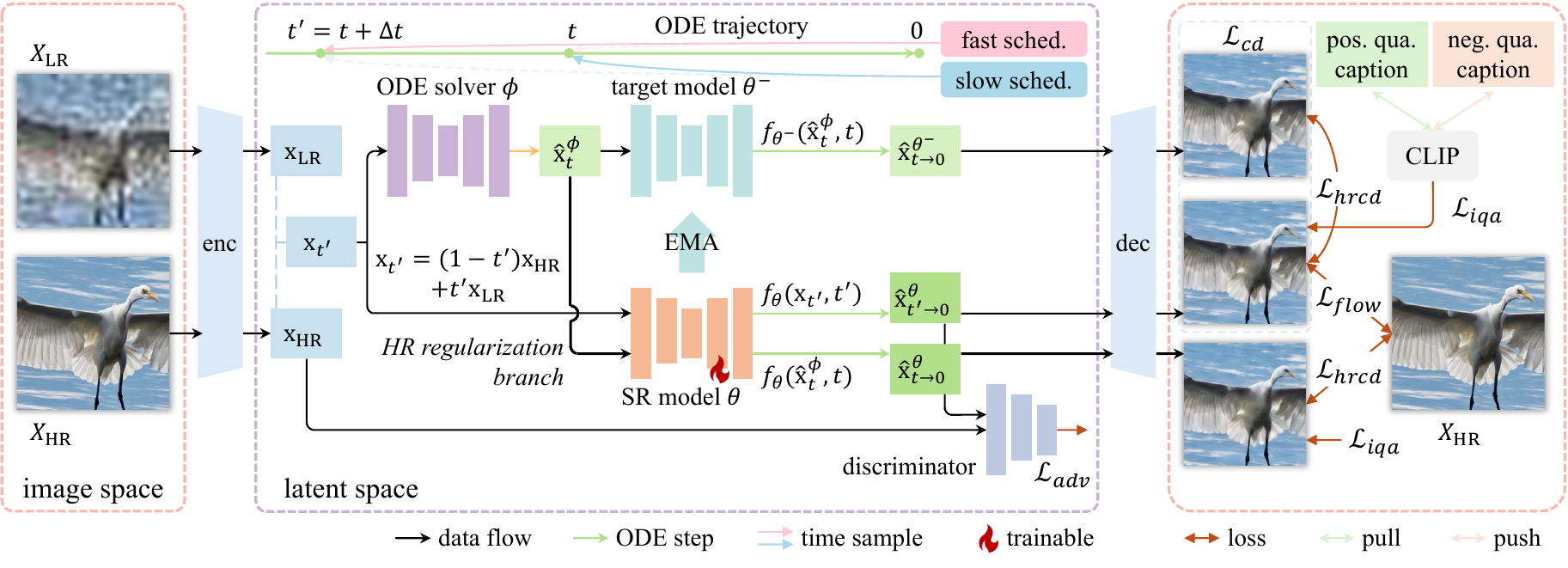}
    \vspace{-2mm}
    \caption
    {Overview of the training process.
    The consistency SR flow distills multi-step, high-quality SR capability into single-step inference, while HR regularization ensures the model converges to the high-resolution target.
    Times $t,t'$ are sampled alternately from fast and slow schedulers.
    Note that the flow loss $\mathcal{L}_{flow}$ and the consistency loss $\mathcal{L}_{hrcd}$ are computed for different samples within each batch.}
    \label{fig:flowsr_overview}
    \vspace{-4mm}
\end{figure*}

\vspace{-4mm}
\paragraph{Single-Step / Few-Step Image SR}
Fast diffusion model-based SR methods require only one or a few sampling steps during inference, typically achieved through knowledge distillation \cite{luhman2021knowledge} or adversarial training \cite{ledig2017photo,xiao2022tackling}.
SinSR~\cite{wang2024sinsr} derives a deterministic sampling process for ResShift and reduces its number of sampling steps to one using distillation.
OSEDiff~\cite{wu2024one} applies variational score distillation~\cite{wang2024prolificdreamer,yin2024one} to enhance one-step restoration quality by minimizing the KL-divergence between the distribution of its generated outputs and that of the pre-trained T2I model.
AddSR \cite{xie2024addsr} achieves 4-step inference by tailoring adversarial diffusion distillation \cite{sauer2024adversarial} for SR.
Similarly, other recent methods, accelerate the inference by using distillation \cite{noroozi2024you,he2024one,dong2025tsd} or GAN \cite{zhang2024degradation,li2024distillation,yue2025arbitrary,chen2025adversarial}.
On the other hand, more efficient SR diffusion processes are constructed using domain shift \cite{cui2024taming} or flow matching \cite{fischer2023boosting}, which allow few-step sampling.
Yet, they still suffer from inefficient modeling designs and suboptimal performance.

\subsection{Consistency Models}
Consistency Models (CMs) \cite{song2023consistency} learn to map any point on the ODE trajectory to its origin, which enable single-step generation and allow trade-offs between quality and computation through multi-step sampling.
CMs can be implemented for modern T2I models in latent space \cite{luo2023latent}.
Consistency Trajectory Models (CTMs) \cite{kim2023consistency} mitigate the multi-step sampling issues in CMs by learning an any-to-any mapping from initial points to final points on the ODE trajectory.
The Phased Consistency Model \cite{wang2024phased} partitions the entire ODE trajectory into multiple sub-trajectory phases and ensures consistency within each phase,
while PeRFlow \cite{yan2024perflow} straightens the sub-trajectories using the Reflow \cite{liu2023instaflow} operation.
Additionally, Consistency Flow Matching \cite{yang2024consistency} further enforces velocity field consistency.

\section{Preliminaries}

\subsection{Rectified Flow}

Rectified flow \cite{liu2022flow} is an ODE-based generative model that constructs a simple, straight trajectory to transform samples between two distributions: $\pi_0$ (\eg, data) and $\pi_1$ (\eg, noise).
Rectified flow defines a linear interpolation path between observed samples $X_0 \sim \pi_0$ and $X_1 \sim \pi_1$ as:
$X_t = tX_1 + (1-t)X_0$ with time $t \in [0, 1]$.
The core idea is to learn a velocity field $v_\theta(X_t, t)$, parameterized by a neural network with weights $\theta$, that matches the derivative of this trajectory.
This is achieved by optimizing the following objective:
$\min\limits_\theta \int_{0}^{1} \mathbb{E}_{X_0 \sim \pi_0, X_1 \sim \pi_1}
[ \| v_\theta(X_t, t) - (X_1 - X_0) \|^2 ] dt$,
which encourages the learned velocity field to align with the direction of the straight path $X_1 - X_0$.
Once trained, samples from $\pi_1$ can be transformed into $\pi_0$ by solving the ODE:
${dX_t} = v_\theta(X_t, t) dt,\ X_1 \sim \pi_1$.
A key advantage of rectified flow lies in its straight trajectories.
By design, the optimal velocity field corresponds to a constant-speed flow between $X_0$ and $X_1$, enabling efficient and stable sampling with only a few ODE steps.

\subsection{Consistency Models}

Consistency models \cite{song2023consistency} are a class of generative models that enable high-quality sample generation with few computational steps.
These models learn to map any point $(X_t,t)$ along a probability flow ODE trajectory directly to its origin $X_\epsilon$, where $\epsilon$ is a fixed small positive number denoting the trajectory's start.
This capability is formalized through the self-consistency property: 
for any pair of points $(X_t, t)$ and $(X_{t'}, t')$ on the same ODE trajectory, the model $f_\theta(\cdot, t)$ is trained to satisfy:
$f_\theta(X_t, t) = f_\theta(X_t', t') = X_\epsilon$ for all $t,t' \in [\epsilon, 1]$.
To ensure consistency during training, the boundary condition $f_\theta(X_\epsilon, \epsilon) = X_\epsilon$ is enforced.
Consistency models eliminate trajectory drift by mapping all intermediate states to a consistent origin, avoiding errors from iterative denoising steps.
This enables them to generate high-fidelity samples even with large step sizes, such as directly mapping from $t=1$ to the origin in a single step.

\section{Methodology}
\label{sec:flowsr_method}

We leverage the generative capability of rectified flow for high-quality SR and explore consistency models to enable fast inference with fewer sampling steps (\ie, one step), without compromising quality.
We define a rectified flow, termed SR flow, for image super-resolution in \cref{sec:flowsr_srflow}.
We then introduce the improved consistency SR flow learning method in \cref{sec:flowsr_consistency} and detail the training in \cref{sec:flowsr_training}.

\subsection{Rectified Flow for SR}
\label{sec:flowsr_srflow}

\paragraph{SR Flow}

We consider Rectified Flow naturally aligns with image super-resolution, where high-resolution (HR) images $X_{\mathrm{HR}} \sim \pi_0$ and their low-resolution (LR) counterparts $X_{\mathrm{LR}} \sim \pi_1$.
We define the forward process as a straight path via a linear interpolation between them:
\begin{equation}
    \label{eq:flowsr_sample_xt}
    X_t = (1 - t) X_{\mathrm{HR}} + t X_{\mathrm{LR}},
\end{equation}
here time $t\in[0, 1]$ with $t=0$ corresponding to $X_\mathrm{HR} \in \mathbb{E}^{H \times W \times 3}$ and $t=1$ corresponding to $X_\mathrm{LR} \in \mathbb{E}^{H \times W \times 3}$, where $X_\mathrm{LR}$ is upscaled to match the spatial dimensions of $X_\mathrm{HR}$.
The learning objective of the SR neural network $v_\theta$ is to regress the conditional vector fields \cite{lipman2022flow,liu2022flow} by following the direction $X_\mathrm{LR} - X_{\mathrm{HR}}$:
\begin{equation}
    \label{eq:flowsr_srflow}
    \mathbb{E}_{t,X_t} \| v_\theta(X_t,t) - (X_{\mathrm{LR}} - X_{\mathrm{HR}}) \|_2^2 .
\end{equation}

Intuitively, this SR flow establishes a smooth transition between HR and LR.
During training, the SR flow model $v_\theta$ implicitly learns to invert degradations (\eg, blur, noise) and recover high-frequency details through intermediate refinements such as edge sharpening and texture synthesis.
At inference, HR images are reconstructed via reverse sampling along the learned trajectory.
Starting from $X_\mathrm{LR}$, we solve the reverse ODE using numerical methods like Euler solver: $X_{t'} = X_t + \Delta t \cdot v_\theta(X_t, t)$ from $t=1$ to $t=0$.
Notably, unlike diffusion-based SR methods that corrupt LR inputs with noise \cite{yue2023resshift,fischer2023boosting,cui2024taming}, our transition process directly starts from LR, which preserves most of the structural information in the LR image, enabling stable and efficient sampling.

\vspace{-4mm}
\paragraph{Faster Inference with SR Flow}
A key advantage of SR flow is its flexibility in sampling steps.
It supports faster inference by reducing the number of sampling steps, requiring as few as a single step that directly maps the LR image to its corresponding HR output:
$\hat{X}_\mathrm{HR}=X_{\mathrm{LR}} - 1 \cdot v_\theta(X_\mathrm{LR}, 1)$.
This is more easily achievable than the noise-to-image mapping or initializing from noisy inputs in ResShift \cite{yue2023resshift}, thanks to the strong correlation between LR and HR images.
Additionally, SR flow also supports any number of sampling steps and produces better visual quality when using iterative multi-step sampling, as illustrated in \cref{fig:flowsr_intro}.

\subsection{Consistency SR Flow}
\label{sec:flowsr_consistency}

We enhance single-step super-resolution in SR flow models by distilling multi-step sampling capability (\eg, four steps) into one via consistency learning (\cref{fig:flowsr_intro}).
To ensure accurate HR reconstruction, we incorporate the HR target explicitly into consistency learning.
Additionally, a fast-slow time sampling schedule is introduced to improve efficiency and robustness.
\cref{fig:flowsr_overview} shows the overall training process.

\begin{figure}[t]
    \centering
    \includegraphics[width=\hsize]{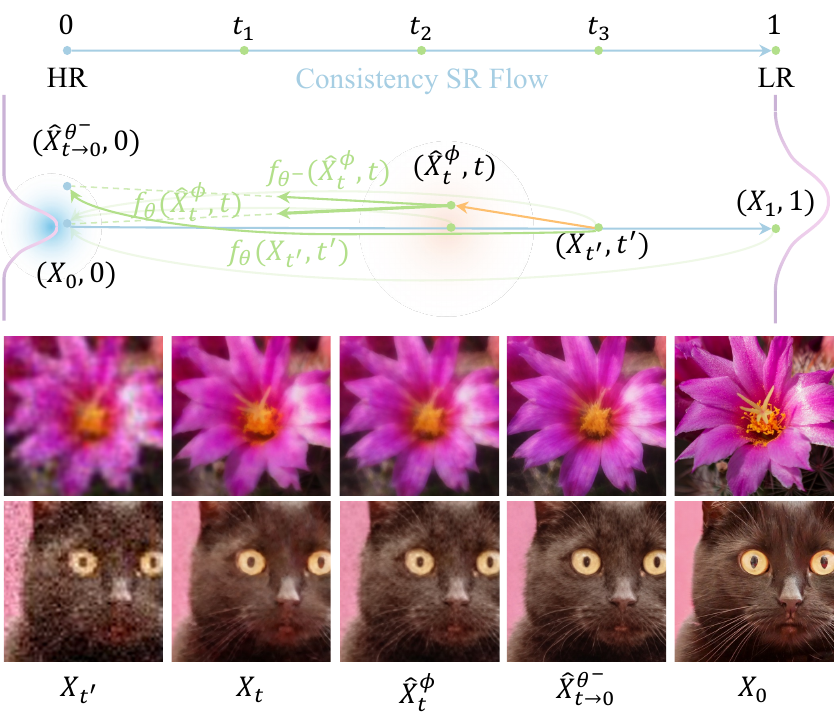}
    \vspace{-6mm}
    \caption
    {Illustration of HR-regularized consistency learning.
    Approximation errors may cause the distillation target $\hat{X}_{t \rightarrow 0}^{\theta^-}$ to deviate from the true high-quality HR target. HR regularization corrects this by enforcing HR alignment under mild perturbations.
    }
    \label{fig:flowsr_ode}
    \vspace{-4mm}
\end{figure}

\vspace{-4mm}
\paragraph{Consistency Distillation in SR Flow}

We start with consistency distillation (CD) to distill the capability of a (pre-trained) teacher SR flow model $v_\phi(X, t)$, enabling high-quality SR in fewer inference steps.
Given timesteps $t$ and $t' = t+\Delta t$ (where $\Delta t > 0$), we first sample $X_{t'}$ from the forward degradation process (\cref{eq:flowsr_sample_xt}).
The teacher model then estimates $\hat{X}_t^\phi$ from its degraded counterpart $X_{t'}$:
\begin{equation}
    \label{eq:flowsr_teacher_step}
    \hat{X}_t^\phi = X_{t'} - \Delta t \cdot v_\phi (X_{t'}, t') ,
\end{equation}
where Euler solver is used.
The consistency function $f_\theta (X_t,t)$ for deriving the origin is defined by:
\begin{equation}
    \label{eq:flowsr_consistency_function}
    f_\theta (X_t,t) = X_t - t \cdot v_\theta(X_t,t) ,
\end{equation}
which maps any degraded input $X_t$ at timestep $t$ to the HR image domain.
The consistency distillation in SR flow learns a vector field $v_\theta$ by minimizing the distance between the prediction of the network $\hat{X}_{t'\rightarrow0}^\theta = f_\theta(X_{t'}, t')$ and the distillation target $\hat{X}_{t\rightarrow0}^{\theta^-} = f_{\theta^-}(\hat{X}_t^\phi, t)$:
\begin{equation}
    \label{eq:flowsr_loss_cd}
    \mathcal{L}_{cd} = \mathbb{E} 
    [ d(f_\theta(X_{t'}, t'), f_{\theta^-}(\hat{X}_t^\phi, t)) ] ,
\end{equation}
where $d(\cdot,\cdot)$ is a distance metric, and $\theta^-$ denotes the target model parameters (\eg, exponential moving average (EMA) of $\theta$).
This distillation loss enables the newly trained model, $f_\theta$, to generate high-quality samples in a single step.

\vspace{-4mm}
\paragraph{HR-Regularized Consistency Learning}

Although the original CD objective theoretically ensures that all trajectory points map to the same origin, it does not explicitly restrict alignment between the generated samples and the corresponding HR targets. This is because velocity field approximation errors from the teacher model $v_\phi$ and the target model $v_{\theta^-}$ can propagate into the distillation target $f_{\theta^-}(\hat{X}_t^\phi, t)$, as shown in \cref{fig:flowsr_ode}.

To address this, we introduce an extra HR regularization into \cref{eq:flowsr_loss_cd}, which directly enforces the student's predictions $\hat{X}_{t\rightarrow0}^{\theta} = f_{\theta}(\hat{X}_t^\phi, t)$ (thus distillation target $f_{\theta^-}(\hat{X}_t^\phi, t)$) to match the ground-truth HR images $X_0$:
\begin{equation}
    \label{eq:flowsr_loss_hrcd}
    \mathcal{L}_{hrcd} = 
    \mathbb{E} [ d(f_\theta(X_{t'}, t'), f_{\theta^-}(\hat{X}_t^\phi, t)) + d(f_{\theta}(\hat{X}_t^\phi, t) , X_0) ] .
\end{equation}

By training with real HR targets, consistency learning reduces reliance on the imperfect teacher and target model.
This dual consistency objective ensures that $f_\theta$  maintains both trajectory consistency and fidelity to HR data.
Note that the input $(\hat{X}_t^\phi, t)$ can be viewed as a perturbed version of sampled $(X_t, t)$ from the forward process in \cref{eq:flowsr_sample_xt}, as visualized in \cref{fig:flowsr_ode}.
This perturbation resembles signal noise and distribution shifts, thereby improving the robustness of the trained model.

\vspace{-4mm}
\paragraph{Fast-Slow Time Scheduling}

To further improve the efficiency and robustness of consistency learning, we propose sampling pairs of timesteps, $t$ and $t'$, from two distinct time schedulers: a ``fast'' scheduler with relatively few timesteps (\eg, 4) and a ``slow'' scheduler with more granular timesteps (\eg, 1000).
The slow scheduler, a standard practice for diffusion or flow models, provides fine-grained learning signals to keep intermediate velocity field predictions well-aligned with the SR flow objectives (\cref{eq:flowsr_srflow}).
In contrast, the fast scheduler emphasizes larger jumps along the trajectory, reflecting more desired one-step inference.

During training, we randomly select one scheduler (fast or slow) to first sample $t+\Delta t$, then sample $t$ in its adjacent region from the other scheduler, and compute $(\hat{X}_t^\phi, t)$ accordingly using \cref{eq:flowsr_teacher_step}, as illustrated in \cref{fig:flowsr_ode}.
This procedure allows flexible and larger jumps $\Delta t$ through the fast sampler and relaxes the requirement for accurately estimating of $X_t$ from $X_{t+\Delta t}$ by running one discretization step of a numerical ODE solver \cite{song2023consistency}.
We hypothesize that introducing mild perturbations and diversity to $\hat{X}_t^\phi$, deviating it from the sampled point $X_t$, enhances the SR model's robustness and mitigates distribution shifts, particularly when combined with HR regularization.

\subsection{Training Consistency SR Flow}
\label{sec:flowsr_training}

\paragraph{SR Flow in Image Space}
We use Stable Diffusion \cite{rombach2022high} as the backbone for our SR model, comprising a VAE to map images $X$ into latent representations $\mathrm{x}$, along with a UNet $\theta$ adapted for velocity field learning.
While the conditional flow matching loss in \cref{eq:flowsr_srflow} can be formulated in latent space, our early experiments observed that such a latent-space objective does not consistently maintain a favorable balance between fidelity and visual quality over varying numbers of inference steps (\eg, from one to four).

In practice, we find that loss in the image space is more effective.
Specifically, during SR flow training, we first predict the velocity prediction $v_\theta(\mathrm{x}_t,t)$ in the latent space and compute the HR latent representation $\hat{\mathrm{x}}_\mathrm{HR}=f_\theta(\mathrm{x}_t,t)$ (see \cref{eq:flowsr_consistency_function}).
Next, we decode $\hat{\mathrm{x}}_\mathrm{HR}$ into the SR image $\hat{X}_\mathrm{HR}$ using the VAE's decoder.
Last, we compare $\hat{X}_\mathrm{HR}$ to the ground-truth HR image $X_\mathrm{HR}$ using an $l_2$ loss and a perceptual (LPIPS \cite{zhang2018unreasonable}) loss:
\begin{equation}
    \label{eq:flowsr_flow}
    \mathcal{L}_{flow} = \|\hat{X}_\mathrm{HR} - X_\mathrm{HR}\|_2^2 + 
    \lambda_{p}\,\text{LPIPS}(\hat{X}_\mathrm{HR}, X_\mathrm{HR}) ,
\end{equation}
where $\lambda_{p}=2$ is set to weight the LPIPS term.

\vspace{-4mm}
\paragraph{Adversarial Loss}
We incorporate an adversarial GAN objective to further enhance SR quality, particularly for texture synthesis.
Specifically, a discriminator $\mathcal{D}$ is trained to distinguish between real HR images and those restored by our model.
We use the pre-trained diffusion model as a feature extractor in the latent space and attach several additional discriminator heads, following \cite{sauer2024fast,yin2024improved,wang2024phased}.
The hinge loss is adopted for training:
\begin{equation}
    \mathcal{L}_{adv} = \mathbb{E}
    \bigl[ \mathrm{max} (0, 1 - \mathcal{D} (\mathrm{x}_\mathrm{HR})) + \mathrm{max} (0, 1 + \mathcal{D} (\hat{\mathrm{x}}_\mathrm{HR})) \bigr] .
\end{equation}
Note that the input to $\mathcal{D}$ is the noised latent with diffusion time $t$ (omitted here for simplicity). The SR model is then trained to minimize this adversarial objective.

\begin{table*}[t]
\caption
{Quantitative comparison with state-of-the-art DM-based SR methods on RealSR \cite{cai2019toward} and DRealSR \cite{wei2020component}.
The best and second-best results are highlighted in \textbf{bold} and {\ul underlined}.}
\label{tab:flowsr_main}
\vspace{-2mm}
\begin{adjustbox}{width=\hsize}
\begin{tabular}{c|c|cccccccccc}
\toprule
Datasets                  & Methods                            & \#Steps                   & PSNR $\uparrow$                     & SSIM $\uparrow$                         & LPIPS $\downarrow$                      & DISTS $\downarrow$                      & FID $\downarrow$                        & NIQE $\downarrow$                     & MUSIQ $\uparrow$                       & MANIQA $\uparrow$                       & CLIPIQA $\uparrow$                      \\ \midrule
                          & StableSR \cite{wang2024exploiting} & 200                       & 24.70                               & 0.7085                                  & 0.3018                                  & 0.2288                                  & 128.51                                  & 5.91                                  & 65.78                                  & 0.6221                                  & 0.6178                                  \\
                          & DiffBIR \cite{lin2024diffbir}      & 50                        & 24.75                               & 0.6567                                  & 0.3636                                  & 0.2312                                  & 128.99                                  & 5.53                                  & 64.98                                  & 0.6246                                  & 0.6463                                  \\
                          & SeeSR \cite{wu2024seesr}           & 50                        & 25.18                               & 0.7216                                  & 0.3009                                  & 0.2223                                  & 125.55                                  & {\ul 5.41}                            & \textbf{69.77}                         & 0.6442                                  & 0.6612                                  \\
                          & PASD \cite{yang2024pixel}          & 20                        & 25.21                               & 0.6798                                  & 0.3380                                  & 0.2260                                  & 124.29                                  & 5.41                                  & 68.75                                  & \textbf{0.6487}                         & 0.6620                                  \\
                          & ResShift \cite{yue2023resshift}    & 15                        & 26.31                               & {\ul 0.7421}                            & 0.3460                                  & 0.2498                                  & 141.71                                  & 7.26                                  & 58.43                                  & 0.5285                                  & 0.5444                                  \\
                          & SinSR \cite{wang2024sinsr}         & 1                         & \textbf{26.28}                      & 0.7347                                  & 0.3188                                  & 0.2353                                  & 135.93                                  & 6.29                                  & 60.80                                  & 0.5385                                  & 0.6122                                  \\
                          & OSEDiff \cite{wu2024one}           & 1                         & 25.15                               & 0.7341                                  & {\ul 0.2921}                            & {\ul 0.2128}                            & {\ul 123.49}                            & 5.65                                  & 69.09                                  & 0.6326                                  & {\ul 0.6693}                            \\
                          & DoSSR \cite{cui2024taming}         & 1                         & {\ul 25.67}                         & 0.7387                                  & 0.3356                                  & 0.2741                                  & 153.25                                  & 11.19                                 & 62.69                                  & 0.5243                                  & 0.6259                                  \\
\multirow{-9}{*}{RealSR}  & \cellcolor[HTML]{EFEFEF}FlowSR     & \cellcolor[HTML]{EFEFEF}1 & \cellcolor[HTML]{EFEFEF}25.54       & \cellcolor[HTML]{EFEFEF}\textbf{0.7434} & \cellcolor[HTML]{EFEFEF}\textbf{0.2728} & \cellcolor[HTML]{EFEFEF}\textbf{0.2013} & \cellcolor[HTML]{EFEFEF}\textbf{112.60} & \cellcolor[HTML]{EFEFEF}\textbf{5.28} & \cellcolor[HTML]{EFEFEF}{\ul 69.22}    & \cellcolor[HTML]{EFEFEF}{\ul 0.6486}    & \cellcolor[HTML]{EFEFEF}\textbf{0.6701} \\ \midrule
                          & StableSR \cite{wang2024exploiting} & 200                       & 28.03                               & 0.7536                                  & 0.3284                                  & 0.2269                                  & 148.98                                  & 6.52                                  & 58.51                                  & 0.5601                                  & 0.6356                                  \\
                          & DiffBIR \cite{lin2024diffbir}      & 50                        & 26.71                               & 0.6571                                  & 0.4557                                  & 0.2748                                  & 166.79                                  & 6.31                                  & 61.07                                  & 0.5930                                  & 0.6395                                  \\
                          & SeeSR \cite{wu2024seesr}           & 50                        & 28.17                               & 0.7691                                  & 0.3189                                  & 0.2315                                  & 147.39                                  & 6.40                                  & {\ul 64.93}                            & 0.6042                                  & 0.6804                                  \\
                          & PASD \cite{yang2024pixel}          & 20                        & 27.36                               & 0.7073                                  & 0.3760                                  & 0.2531                                  & 156.13                                  & \textbf{5.55}                         & 64.87                                  & {\ul 0.6169}                            & 0.6808                                  \\
                          & ResShift \cite{yue2023resshift}    & 15                        & 28.46                               & 0.7673                                  & 0.4006                                  & 0.2656                                  & 172.26                                  & 8.12                                  & 50.60                                  & 0.4586                                  & 0.5342                                  \\
                          & SinSR \cite{wang2024sinsr}         & 1                         & 28.36                               & 0.7515                                  & 0.3665                                  & 0.2485                                  & 170.57                                  & 6.99                                  & 55.33                                  & 0.4884                                  & 0.6383                                  \\
                          & OSEDiff \cite{wu2024one}           & 1                         & 27.92                               & 0.7835                                  & \textbf{0.2968}                         & {\ul 0.2165}                            & {\ul 135.30}                            & 6.49                                  & 64.65                                  & 0.5899                                  & {\ul 0.6963}                            \\
                          & DoSSR \cite{cui2024taming}         & 1                         & \textbf{28.55}                      & \textbf{0.7991}                         & 0.3353                                  & 0.2801                                  & 166.19                                  & 12.25                                 & 56.72                                  & 0.4623                                  & 0.5739                                  \\
\multirow{-9}{*}{DRealSR} & \cellcolor[HTML]{EFEFEF}FlowSR     & \cellcolor[HTML]{EFEFEF}1 & \cellcolor[HTML]{EFEFEF}{\ul 28.50} & \cellcolor[HTML]{EFEFEF}{\ul 0.7859}    & \cellcolor[HTML]{EFEFEF}{\ul 0.2975}    & \cellcolor[HTML]{EFEFEF}\textbf{0.2115} & \cellcolor[HTML]{EFEFEF}\textbf{130.30} & \cellcolor[HTML]{EFEFEF}{\ul 6.13}    & \cellcolor[HTML]{EFEFEF}\textbf{65.46} & \cellcolor[HTML]{EFEFEF}\textbf{0.6172} & \cellcolor[HTML]{EFEFEF}\textbf{0.7074} \\ \bottomrule
\end{tabular}
\end{adjustbox}
\vspace{-4mm}
\end{table*}

\vspace{-4mm}
\paragraph{Image Quality Alignment Loss}

To further align the SR outputs with desirable image quality attributes (\eg, high-resolution, sharp, detailed), we propose an image quality alignment loss based on a text-image contrastive loss.
Specifically, we use large vision-language models to generate a positive quality caption $c_{pos}$ (\eg, ``image quality is good, with clear details and vibrant colors'') from the HR image and a negative quality caption $c_{neg}$ (\eg, ''image quality is poor, blurry, and noisy'') from its LR counterpart.
We then encode these captions using the text encoder $\mathcal{E}_T$ of CLIP \cite{radford2021learning}, while the SR result $\hat{X}$ is encoded via CLIP’s image encoder $\mathcal{E}_I$ (ViT-L/14 is used).
Image quality alignment loss encourages $\hat{X}$ to be close to $c_{pos}$ and far from $c_{neg}$:
\begin{equation}
    \mathcal{L}_{iqa} = - \mathrm{log} \frac {\mathrm{exp}(\mathrm{sim}(\hat{X}, c_{pos}))}
    {\mathrm{exp}(\mathrm{sim}(\hat{X}, c_{pos})) + \mathrm{exp}(\mathrm{sim}(\hat{X}, c_{neg}))} ,
\end{equation}
where $\mathrm{sim}(X,c) = \mathrm{cos}(\mathcal{E}_I(X), \mathcal{E}_T(c))$.
Intuitively, this loss implicitly guides the network to improve perceptual fidelity and clarity by bridging the gap between human-perceived ``good'' and ``bad'' image attributes.

\vspace{-4mm}
\paragraph{Putting Things Together}

We first pre-train the SR flow model using the basic conditional flow loss in \cref{eq:flowsr_flow}.
Then we continue with the consistency SR flow training by fine-tuning it.
The overall loss function is defined as:
\begin{equation}
    \label{eq:flowsr_loss}
    \mathcal{L} = \mathcal{L}_{flow}
    + \lambda_{cd} \mathcal{L}_{hrcd}
    + \lambda_{adv} \mathcal{L}_{adv}
    + \lambda_{iqa} \mathcal{L}_{iqa} ,
\end{equation}
where $\lambda_{cd}$, $\lambda_{adv}$, and $\lambda_{iqa}$ are hyperparameters.

\section{Experiments}
\label{sec:flowsr_exp}

\subsection{Experimental Settings}

\paragraph{Implementation Details}
Our model is based on the pretrained SD 2.1-base~\cite{rombach2022high}.
During training, we fine-tune only the U-Net using a LoRA~\cite{hu2022lora} rank of 32.
The learning rate is set to 2e-5,
with a training patch size of $512 \times 512$ and a batch size of 16.
The training process consists of two stages: we firstly train the SR flow model for 10k iterations, followed by consistency learning for an additional 20k iterations.
The loss weights $\lambda_{cd},\lambda_{adv},\lambda_{iqa}$ are set to 0.1, 0.05, and 0.1, respectively.

\vspace{-4mm}
\paragraph{Data}
We train our model using LSDIR~\cite{li2023lsdir} and the first 10K face images from FFHQ~\cite{karras2019style}.
The LR-HR training pairs are synthesized using the degradation pipeline of Real-ESRGAN~\cite{wang2021real}.
To generate image quality captions for HR and LR images, we use Qwen2-VL~\cite{wang2024qwen2}.
We adopt the real-world test set of StableSR~\cite{wang2024exploiting} for evaluation and comparison. The test sets include RealSR \cite{cai2019toward} and DRealSR \cite{wei2020component}.

\vspace{-4mm}
\paragraph{Compared Methods}
We compare our method against several state-of-the-art diffusion-based SR approaches,
including multi-step StableSR \cite{wang2024exploiting}, DiffBIR \cite{lin2024diffbir}, SeeSR \cite{wu2024seesr}, PASD \cite{yang2024pixel}, ResShift \cite{yue2023resshift}, and one-step SinSR \cite{wang2024sinsr}, OSEDiff \cite{wu2024one}, and DoSSR \cite{cui2024taming}.

\vspace{-4mm}
\paragraph{Evaluation Metrics}
We evaluate our model using both reference-based and no-reference metrics.
For fidelity, we report PSNR and SSIM (on the Y channel in YCbCr space), and for perceptual quality, we use LPIPS~\cite{zhang2018unreasonable} and DISTS~\cite{ding2020image}.
FID~\cite{heusel2017gans} is reported to compare the distribution of restored images with the ground truth.
For no-reference evaluation, we use NIQE \cite{mittal2012making}, MUSIQ~\cite{ke2021musiq}, MANIQA~\cite{yang2022maniqa}, and CLIPIQA~\cite{wang2023exploring}.

\begin{figure*}[t]
    \centering
    \includegraphics[width=\hsize]{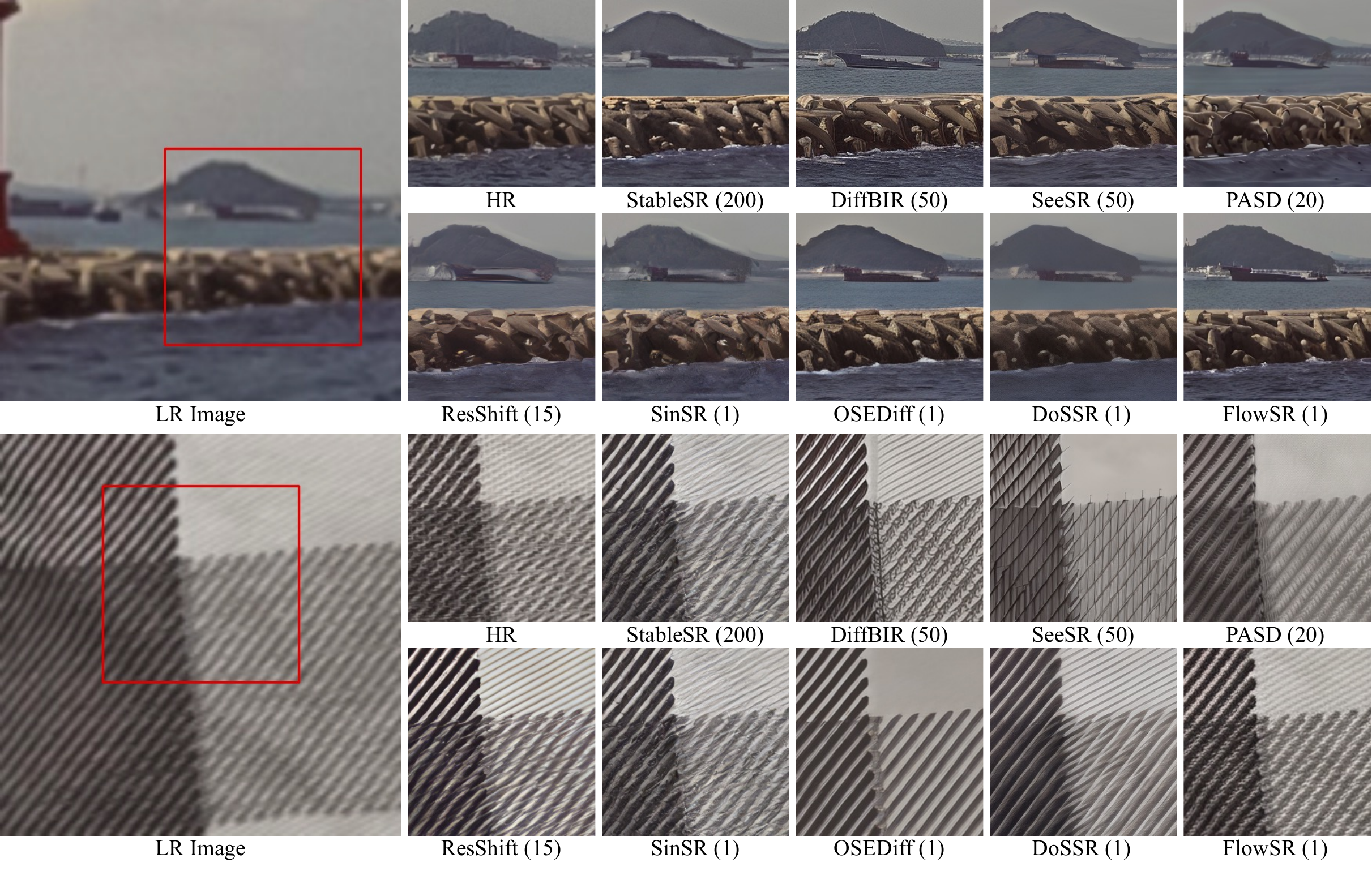}
    \vspace{-8mm}
    \caption
    {Visual comparisons of different SR methods on real-world examples. 
    The number of sampling steps are indicated in parentheses. Please zoom in for a better view.}
    \label{fig:flowsr_vis_main}
    \vspace{-4mm}
\end{figure*}

\subsection{Comparison with State-of-the-Art Methods}

\paragraph{Quantitative Comparisons}
\Cref{tab:flowsr_main} shows that our approach achieves superior or at least competitive performance compared to recent state-of-the-art methods.
Compared with multi-step diffusion-based methods such as SeeSR \cite{wu2024seesr} and PASD \cite{yang2024pixel}, our FlowSR obtains equivalent or better no-reference metrics, such as MUSIQ, MANIQA, and CLIPIQA, and surpasses these methods in fidelity-based scores, \eg PSNR and LPIPS, while reducing inference to a single step.
Compared to efficient diffusion-based SR methods such as ResShift \cite{yue2023resshift}, our method demonstrates a clear advantage.
Against the strong DDPM-based one-step competitor OSEDiff~\cite{wu2024one}, FlowSR achieves superior performance across nearly all metrics.
Overall, these results demonstrate that our flow-based SR method effectively balances fidelity and perceived quality while showing superiority over compared methods.

\vspace{-4mm}
\paragraph{Qualitative Comparisons}

\cref{fig:flowsr_vis_main} shows visual comparisons of two real-world examples.
Multi-step DM-based methods, such as DiffBIR and PASD, often generate rich but inaccurate textures, while one-step methods like SinSR and OSEDiff tend to produce blurry or less detailed results.
In contrast, our method generates faithful SR results, effectively recovering structures such as the stone breakwater and the textures of the cloth.

\subsection{Ablation Studies}

\paragraph{Effects of SR Flow}

\begin{figure}[t]
    \centering
    \captionsetup[subfigure]{font=scriptsize,labelformat=empty,justification=centering}

    \begin{subfigure}{0.24\hsize}
        \includegraphics[width=\hsize]{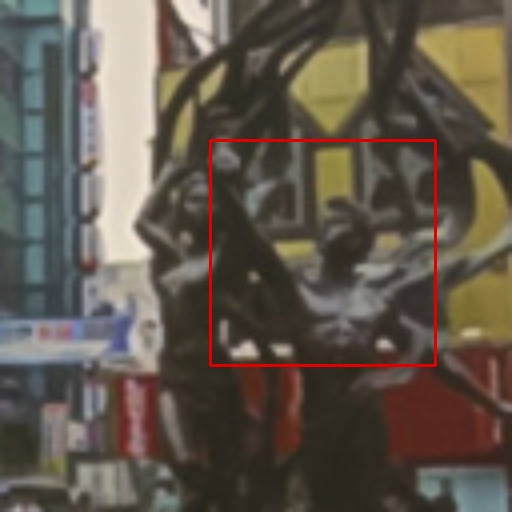}
        \caption{LR Image}
    \end{subfigure}
    \begin{subfigure}{0.24\hsize}
        \includegraphics[width=\hsize]{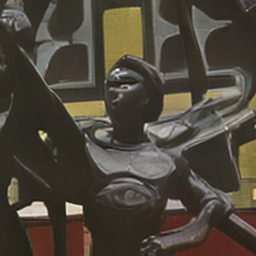}
        \caption{Noise$\rightarrow$HR (4)}
    \end{subfigure}
    \begin{subfigure}{0.24\hsize}
        \includegraphics[width=\hsize]{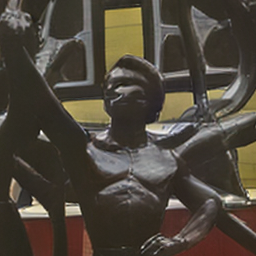}
        \caption{noised LR$\rightarrow$HR (4)}
    \end{subfigure}
    \begin{subfigure}{0.24\hsize}
        \includegraphics[width=\hsize]{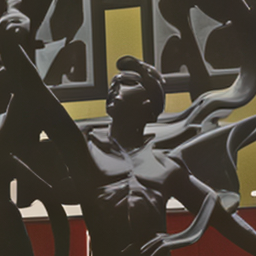}
        \caption{SR Flow (4)}
    \end{subfigure}
    
    \begin{subfigure}{0.24\hsize}
        \includegraphics[width=\hsize]{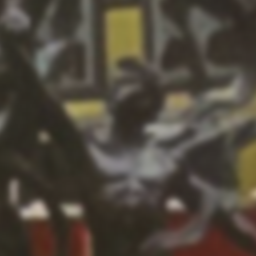}
        \caption{Zoomed LR}
    \end{subfigure}
    \begin{subfigure}{0.24\hsize}
        \includegraphics[width=\hsize]{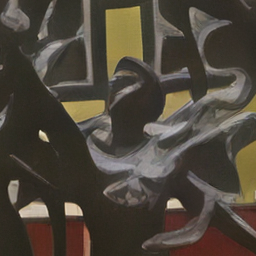}
        \caption{Noise$\rightarrow$HR (1)}
    \end{subfigure}
    \begin{subfigure}{0.24\hsize}
        \includegraphics[width=\hsize]{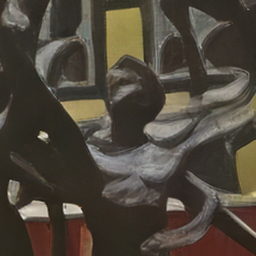}
        \caption{noised LR$\rightarrow$HR (1)}
    \end{subfigure}
    \begin{subfigure}{0.24\hsize}
        \includegraphics[width=\hsize]{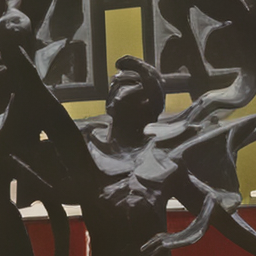}
        \caption{SR Flow (1)}
    \end{subfigure}

    \vspace{-2mm}
    \caption
    {Ablation of SR flow.
    Our SR flow model produces more accurate SR results with improved structure with more steps.}
    \label{fig:flowsr_vis_ablation_flow}
    \vspace{-4mm}
\end{figure}

\begin{table}[t]
\centering
\vspace{-2mm}
\caption
{Ablation of SR flow.
SR flow mapping LR to HR outperforms alternative flow formulations that start from noise or ResShift \cite{yue2023resshift}-style noised LR on RealSR and DRealSR.}
\label{tab:flowsr_ablation_flow}
\begin{adjustbox}{width=\hsize}
\begin{tabular}{llccccc}
\toprule
\multirow{2}{*}{\#Steps} & \multirow{2}{*}{Method}  & \multicolumn{2}{c}{RealSR}         &           & \multicolumn{2}{c}{DRealSR}        \\ \cmidrule{3-4} \cmidrule{6-7} 
                         &                          & PSNR $\uparrow$ & MUSIQ $\uparrow$ &           & PSNR $\uparrow$ & MUSIQ $\uparrow$ \\ \midrule
\multirow{3}{*}{4}       & Noise$\rightarrow$HR     & 23.83           & 66.01            &           & 26.75           & 62.76            \\
                         & noised LR$\rightarrow$HR & 24.47           & 65.15            &           & 27.74           & 61.76            \\
                         & SR Flow (ours)           & \textbf{25.00}  & \textbf{68.09}   &           & \textbf{27.98}  & \textbf{65.05}   \\ \midrule
\multirow{3}{*}{1}       & Noise$\rightarrow$HR     & 24.82           & 65.17            &           & 28.07           & 59.34            \\
                         & noised LR$\rightarrow$HR & 24.67           & 65.25            &           & 28.03           & 60.64            \\
                         & SR Flow (ours)           & \textbf{25.51}  & \textbf{67.40}   & \textbf{} & \textbf{28.65}  & \textbf{62.53}   \\
\bottomrule
\end{tabular}
\end{adjustbox}
\vspace{-4mm}
\end{table}

We first study the flow formulation for image super-resolution.
Our SR flow is a straight path between $\pi_0 = X_{\mathrm{HR}}$ and $\pi_1 = X_{\mathrm{LR}}$.
We consider alternative formulations starting from noise, where $\pi_1 = \mathcal{N}(0, 1)$, as used in pre-trained T2I models, or starting from noised LR, where $\pi_1 = X_\mathrm{LR} + \mu \epsilon$, with $\epsilon \sim \mathcal{N}(0, 1)$, as proposed in ResShift \cite{yue2023resshift}.
For these alternatives, we concatenate $X_t$ with $X_\mathrm{LR}$ as the input to provide the LR condition.
The trained models are evaluated on the RealSR \cite{cai2019toward} and DRealSR \cite{wei2020component} datasets, as shown in \Cref{tab:flowsr_ablation_flow}.
We observe that with more inference steps, the image quality (MUSIQ) generally improves, while the fidelity (PSNR) slightly decreases.
\cref{fig:flowsr_vis_ablation_flow} presents the visual comparisons, where our SR flow model produces more accurate predictions, with further structure and overall quality enhancement as additional sampling steps (\ie, 4 steps) are applied.
Notably, these results demonstrate that the model trained using our SR flow clearly outperforms other transition variants in both single-step and multi-step inference settings.

\begin{figure}[t]
    \centering
    \captionsetup[subfigure]{font=scriptsize,labelformat=empty,justification=centering}

    \begin{subfigure}{0.24\hsize}
        \includegraphics[width=\hsize]{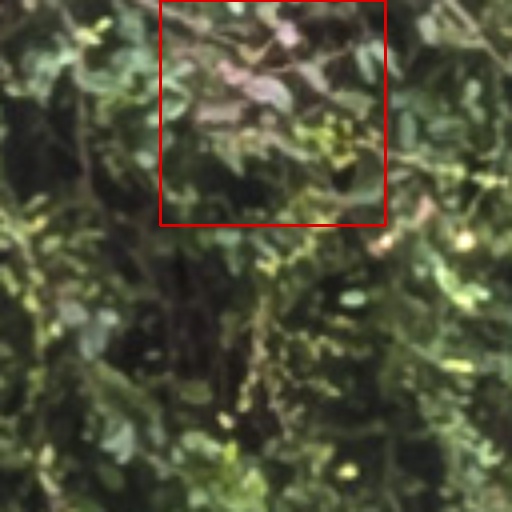}
    \end{subfigure}
    \begin{subfigure}{0.24\hsize}
        \includegraphics[width=\hsize]{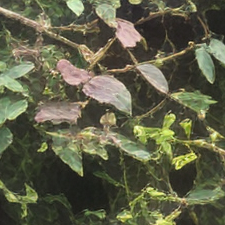}
    \end{subfigure}
    \begin{subfigure}{0.24\hsize}
        \includegraphics[width=\hsize]{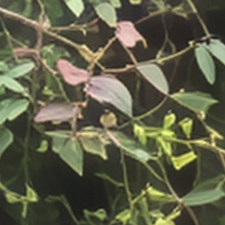}
    \end{subfigure}
    \begin{subfigure}{0.24\hsize}
        \includegraphics[width=\hsize]{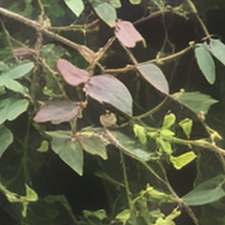}
    \end{subfigure}
    
    \begin{subfigure}{0.24\hsize}
        \includegraphics[width=\hsize]{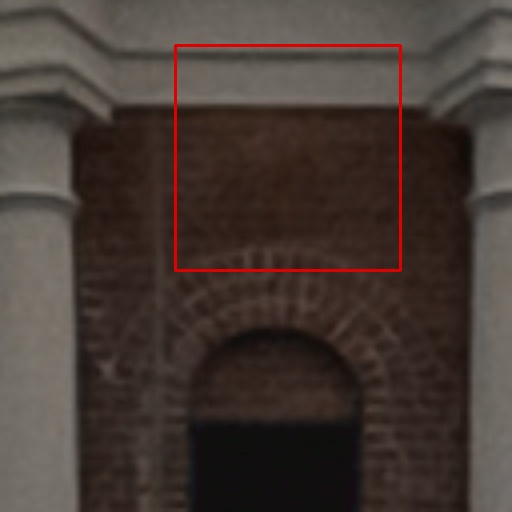}
        \caption{LR Image}
    \end{subfigure}
    \begin{subfigure}{0.24\hsize}
        \includegraphics[width=\hsize]{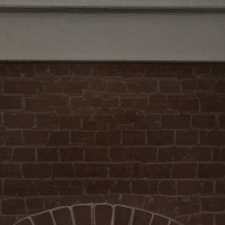}
        \caption{SR Flow}
    \end{subfigure}
    \begin{subfigure}{0.24\hsize}
        \includegraphics[width=\hsize]{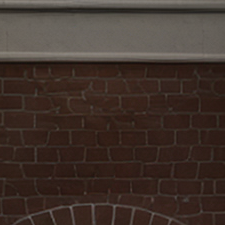}
        \caption{w/ $\mathcal{L}_{cd}$}
    \end{subfigure}
    \begin{subfigure}{0.24\hsize}
        \includegraphics[width=\hsize]{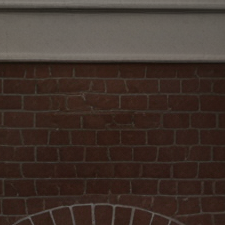}
        \caption{w/ $\mathcal{L}_{hrcd}$}
    \end{subfigure}

    \vspace{-2mm}
    \caption
    {Ablation of consistency learning.
    Our HR-regularized consistency learning effectively reduces distortions in SR outputs while producing high-quality and sharper results (zoom in).}
    \label{fig:flowsr_vis_ablation_consistency}
    \vspace{-2mm}
\end{figure}

\vspace{-4mm}
\paragraph{Effects of consistency learning}
Next, we fine-tune the pre-trained SR flow model using the consistency objective described in \cref{sec:flowsr_consistency}.
As shown in \Cref{tab:flowsr_ablation_consistency}, enforcing self-consistency $\mathcal{L}_{cd}$ across neighboring time steps stabilizes intermediate representations and often reduces LPIPS.
However, when the teacher’s predictions deviate from the true HR manifold, these errors may propagate to the student model, leading to diminished IQA scores.
We address this by introducing the HR regularization term in \cref{eq:flowsr_loss_hrcd}, which directly aligns student predictions $f_{\theta}(\hat{X}_t^\phi, t)$ with real HR images, mitigating the teacher’s approximation errors.
This extra ground-truth constraint helps the model recover fine textural details, which boosts IQA metrics (\eg, MUSIQ and MANIQA) while preserves perceptual fidelity.
Consistency learning enhances the robustness of flow-based inference by distilling multi-step capabilities into a single step, allowing for faster sampling without sacrificing quality.
Additionally, HR regularization alone (w/ $\mathcal{L}_{hr}$) provides minimal IQA gains. This highlights the importance of consistency learning and its synergy with HR regularization for acceleration in SR flow.
\cref{fig:flowsr_vis_ablation_consistency} presents visual comparisons, showing that consistency learning helps resolve distorted textures in the baseline SR flow model.
Notably, HR-regularized consistency learning generates more realistic and sharper results compared to its baseline.

\begin{table}[t]
\caption
{Ablation of consistency learning.
Fine-tuned SR flow models with consistency objectives are evaluated on DRealSR.}
\label{tab:flowsr_ablation_consistency}
\vspace{-2mm}
\begin{adjustbox}{width=\hsize}
\begin{tabular}{llccccc}
\toprule
\#Steps            & Method                  & \multicolumn{1}{l}{PSNR $\uparrow$} & \multicolumn{1}{l}{LPIPS $\downarrow$} & \multicolumn{1}{l}{MUSIQ $\uparrow$} & \multicolumn{1}{l}{MANIQA $\uparrow$} & \multicolumn{1}{l}{CLIPIQA $\uparrow$} \\ \midrule
\multirow{3}{*}{4} & SR Flow                 & 27.98                               & 0.2763                                 & 65.05                                & 0.6296                                & 0.6681                                 \\
                   & w/ $\mathcal{L}_{cd}$   & 28.64                               & 0.2654                                 & 61.82                                & 0.6042                                & 0.6180                                 \\
                   & w/ $\mathcal{L}_{hrcd}$ & 28.44                               & 0.2799                                 & 65.71                                & 0.6374                                & 0.6709                                 \\ \midrule
\multirow{4}{*}{1} & SR Flow                 & 28.65                               & 0.2821                                 & 62.53                                & 0.5895                                & 0.6545                                 \\
                   & w/ $\mathcal{L}_{cd}$   & 28.59                               & 0.2609                                 & 61.69                                & 0.5949                                & 0.6391                                 \\
                   & w/ $\mathcal{L}_{hr}$   & 28.85                               & 0.2805                                 & 62.46                                & 0.5871                                & 0.6589                                 \\
                   & w/ $\mathcal{L}_{hrcd}$ & 28.62                               & 0.2830                                 & 64.77                                & 0.6156                                & 0.6748                                 \\ \bottomrule
\end{tabular}
\end{adjustbox}
\vspace{-4mm}
\end{table}

\vspace{-4mm}
\paragraph{Analysis of Fast-Slow Time Sampling}
We evaluate the effectiveness of fast-slow time sampling by comparing it to the $N$-interval scheduler \cite{song2023consistency} and by varying the number of fast-scheduler timesteps, while fixing the slow scheduler fixed at 1000 timesteps.
When the fast scheduler uses only one timestep, we set $t+\Delta t=1$ and then sample $t$ from the slow scheduler (see \cref{eq:flowsr_teacher_step}).
As shown in \Cref{tab:flowsr_ablation_fastslow}, increasing $\Delta t$ (thus reducing the number of intervals) within a reasonable range improves IQA metrics.
A plausible explanation is that the slightly larger perturbations introduced at intermediate predictions $\hat{X}_t^\phi$ from $X_t$ encourage the model to be more robust to distribution shifts; see \cref{fig:flowsr_ode}.
This, in turn, reduces variance in its estimates of the ODE trajectory's starting point, \ie, HR, resulting in sharper restorations.
Moreover, compared to the $N$-interval strategy or slow-only scheduling, our fast-slow scheduler further boosts IQA metrics while maintaining comparable fidelity.
This is because the slow scheduler ensures fine-grained SR flow estimation, whereas the fast scheduler enables efficient sampling.
Based on these findings, we set the fast scheduler to 4 timesteps throughout our experiments.

\begin{figure}[t]
    \centering
    \captionsetup[subfigure]{font=scriptsize,labelformat=empty,justification=centering}

    \begin{subfigure}{0.19\hsize}
        \includegraphics[width=\hsize]{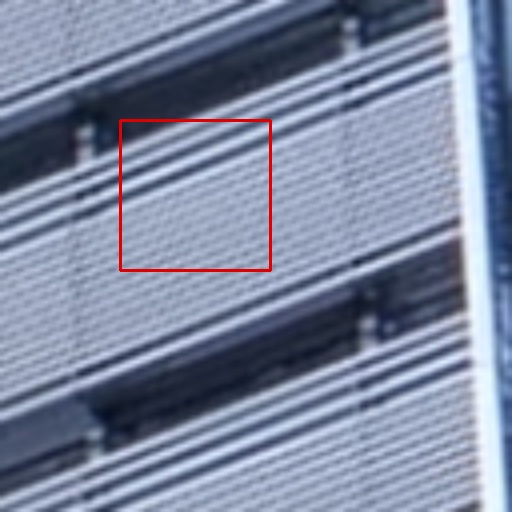}
        \caption{LR Image}
    \end{subfigure}
    \begin{subfigure}{0.19\hsize}
        \includegraphics[width=\hsize]{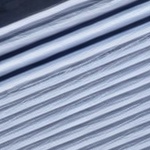}
        \caption{w/o $\mathcal{L}_{hrcd}$}
    \end{subfigure}
    \begin{subfigure}{0.19\hsize}
        \includegraphics[width=\hsize]{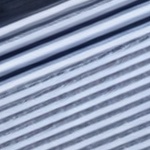}
        \caption{w/o $\mathcal{L}_{adv}$}
    \end{subfigure}
    \begin{subfigure}{0.19\hsize}
        \includegraphics[width=\hsize]{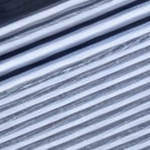}
        \caption{w/o $\mathcal{L}_{iqa}$}
    \end{subfigure}
    \begin{subfigure}{0.19\hsize}
        \includegraphics[width=\hsize]{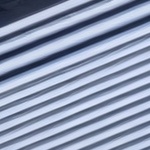}
        \caption{Ours}
    \end{subfigure}

    \vspace{-2mm}
    \caption
    {Ablation of training loss.}
    \label{fig:flowsr_vis_ablation_loss}
    \vspace{-2mm}
\end{figure}

\begin{table}[t]
\caption
{Ablation of fast-slow time sampling.
We compare our fast-slow sampling approach with the $N$-interval method on DRealSR.
The number of $N$ and fast-scheduler timesteps are indicated in parentheses.}
\label{tab:flowsr_ablation_fastslow}
\vspace{-2mm}
\begin{adjustbox}{width=\hsize}
\begin{tabular}{lccccc}
\toprule
Method            & \multicolumn{1}{l}{PSNR $\uparrow$} & \multicolumn{1}{l}{LPIPS $\downarrow$} & \multicolumn{1}{l}{MUSIQ $\uparrow$} & \multicolumn{1}{l}{MANIQA $\uparrow$} & \multicolumn{1}{l}{CLIPIQA $\uparrow$} \\ \midrule
$N$-Interval (50) & 28.61                               & 0.2807                                 & 62.48                                & 0.5914                                & 0.6599                                 \\
$N$-Interval (18) & 28.74                               & 0.2792                                 & 62.90                                & 0.5972                                & 0.6605                                 \\
$N$-Interval (4)  & 28.66                               & 0.2818                                 & 63.03                                & 0.5972                                & 0.6617                                 \\
Slow Only (1000)  & 28.78                               & 0.2801                                 & 62.72                                & 0.5916                                & 0.6586                                 \\
Fast-Slow (8)     & 28.54                               & 0.2760                                 & 64.01                                & 0.6098                                & 0.6634                                 \\
Fast-Slow (4)     & 28.62                               & 0.2830                                 & 64.77                                & 0.6156                                & 0.6748                                 \\
Fast-Slow (1)     & 28.79                               & 0.2835                                 & 63.01                                & 0.5948                                & 0.6605                                 \\ \bottomrule
\end{tabular}
\end{adjustbox}
\vspace{-2mm}
\end{table}

\begin{table}[t]
\caption
{Ablation of training loss on DRealSR.}
\label{tab:flowsr_ablation_loss}
\vspace{-2mm}
\begin{adjustbox}{width=\hsize}
\begin{tabular}{lccccc}
\toprule
Method                   & PSNR $\uparrow$ & LPIPS $\downarrow$ & MUSIQ $\uparrow$ & MANIQA $\uparrow$ & CLIPIQA $\uparrow$ \\ \midrule
w/o $\mathcal{L}_{hrcd}$ & 28.52           & 0.3013             & 64.42            & 0.6061            & 0.6954             \\
w/o $\mathcal{L}_{adv}$  & 28.45           & 0.2961             & 65.29            & 0.6154            & 0.7099             \\
w/o $\mathcal{L}_{iqa}$  & 28.64           & 0.2958             & 64.83            & 0.6111            & 0.7012             \\
Ours                     & 28.50           & 0.2975             & 65.46            & 0.6172            & 0.7074             \\ \bottomrule
\end{tabular}
\end{adjustbox}
\vspace{-4mm}
\end{table}

\vspace{-4mm}
\paragraph{Influence of loss function}
In addition to flow and consistency learning, we incorporate a GAN loss and an image quality alignment loss to further enhance SR performance, as detailed in \cref{eq:flowsr_loss}.
As demonstrated in \Cref{tab:flowsr_ablation_loss} and \cref{fig:flowsr_vis_ablation_loss}, these losses all improve no-reference metrics and visual quality.
The ablation highlights that eliminating any of these losses results in reduced SR quality, with consistency learning contributing a substantial improvement.

\section{Conclusion}

This paper presents FlowSR, a new approach for efficient one-step image super-resolution.
FlowSR reformulates SR as a rectified flow, leveraging the strengths of iterative generative modeling.
To enable high-quality single-step inference, we incorporate consistency learning and devise HR regularization to address its distillation target drifting issue.
Additionally, a fast-slow time scheduling strategy is designed to enhance the efficiency and robustness of the consistency SR flow model.
FlowSR contributes to the advancement of efficient real-world SR applications.

\section*{Acknowledgement}

The work described in this paper was supported in part by the National Key R\&D Program of China (Grant No. 2023YFE0202700),
Research Grants Council of the Hong Kong Special Administrative Region, China, under Project 14200824;
and by the Hong Kong Innovation and Technology Fund, under Project MHP/092/22.

{
    \small
    \bibliographystyle{ieeenat_fullname}
    \bibliography{ref}
}

\begin{table*}[t]
\caption
{Quantitative comparisons of different methods on the DIV2K-Val dataset.}
\label{tab:flowsr_div2k}
\begin{adjustbox}{width=\hsize}
\begin{tabular}{ccccccccccc}
\toprule
Methods                            & \#Steps & PSNR $\uparrow$ & SSIM $\uparrow$ & LPIPS $\downarrow$ & DISTS $\downarrow$ & FID $\downarrow$ & NIQE $\downarrow$ & MUSIQ $\uparrow$ & MANIQA $\uparrow$ & CLIPIQA $\uparrow$ \\ \midrule
StableSR \cite{wang2024exploiting} & 200     & 23.26           & 0.5726          & 0.3113             & 0.2048             & \textbf{24.44}   & 4.76              & 65.92            & 0.6192            & 0.6771             \\
DiffBIR \cite{lin2024diffbir}      & 50      & 23.64           & 0.5647          & 0.3524             & 0.2128             & 30.72            & 4.70              & 65.81            & 0.6210            & 0.6704             \\
SeeSR \cite{wu2024seesr}           & 50      & 23.68           & 0.6043          & 0.3194             & {\ul 0.1968}       & 25.90            & 4.81              & {\ul 68.67}      & {\ul 0.6240}      & \textbf{0.6936}    \\
PASD \cite{yang2024pixel}          & 20      & 23.14           & 0.5505          & 0.3571             & 0.2207             & 29.20            & \textbf{4.36}     & \textbf{68.95}   & \textbf{0.6483}   & 0.6788             \\
ResShift \cite{yue2023resshift}    & 15     & \textbf{24.65}  & 0.6181          & 0.3349             & 0.2213             & 36.11            & 6.82              & 61.09            & 0.5454            & 0.6071             \\
SinSR \cite{wang2024sinsr}         & 1      & 24.41           & 0.6018          & 0.3240             & 0.2066             & 35.57            & 6.02              & 62.82            & 0.5386            & 0.6471             \\
OSEDiff \cite{wu2024one}           & 1      & 23.72           & 0.6108          & {\ul 0.2941}       & 0.1976             & 26.32            & 4.71              & 67.97            & 0.6148            & 0.6683             \\
DoSSR \cite{cui2024taming}         & 1      & 24.35           & \textbf{0.6265} & 0.3725             & 0.2786             & 50.27            & 10.38             & 58.44            & 0.5024            & 0.6187             \\
\rowcolor[HTML]{EFEFEF} 
FlowSR                             & 1      & {\ul 24.42}     & {\ul 0.6192}    & \textbf{0.2798}    & \textbf{0.1847}    & {\ul 24.52}      & {\ul 4.63}        & 68.22            & 0.6193            & {\ul 0.6901}       \\ \bottomrule
\end{tabular}
\end{adjustbox}
\end{table*}

In this supplementary material, we first provide additional details about our FlowSR in \cref{sec:flowsr_supp_method}.
Next, we present more experimental results in \cref{sec:flowsr_supp_exp}.
Finally, we discuss the limitations of our approach and outline potential future directions in \cref{sec:flowsr_supp_limitation}.

\section{Implementation Details}
\label{sec:flowsr_supp_method}

We first fine-tune the pre-trained SD model \cite{rombach2022high} to adapt it to our SR flow learning objectives.
The fine-tuned SR flow model is then used to initialize both the SR model $\theta$ and the teacher model $\phi$.
A default text prompt is used for the SD model.
During consistency SR flow training, each training batch is split into two groups: one for SR flow learning and the other for consistency learning.
This approach ensures that the fine-tuned SR model still learns accurate SR flow while also acquiring distilled one-step high-quality inference capability.

For the fast-slow time scheduling, the adjacent time steps $t$ and $t' = t + \Delta t$ are sampled as follows: we first randomly select either the fast scheduler or the slow scheduler and use it to sample $t'$. Then, the other scheduler is used to sample $t$.
If the fast scheduler is chosen first, $t$ is sampled from the range between $t'$ and its predecessor timestep. Conversely, if the slow scheduler is chosen first, $t$ is sampled from the next time point less than $t'$.
This approach ensures that the jump $\Delta t$ remains flexible.

We also observe that the choice of timestep shifting and sampling plays a crucial role in SR flow learning, and we provide an ablation study in \cref{subsec:flowsr_supp_timestep} to further analyze this.

\section{More Results}
\label{sec:flowsr_supp_exp}

\subsection{Evaluation on DIV2K-Val}

We also evaluate our method on the DIV2K-Val dataset \cite{agustsson2017ntire,wang2024exploiting}.
\Cref{tab:flowsr_div2k} provides a quantitative comparison of various SR methods.
Across all reference-based metrics, our FlowSR achieves state-of-the-art performance or performs on par with the best existing methods.
For no-reference metrics, while FlowSR performs worse than the multi-step SD-based PASD \cite{yang2024pixel}, it remains the best-performing model among all single-step sampling methods.
These results demonstrate the effectiveness and superiority of our method.

\subsection{Model efficiency}

We present the model parameters, MACs, and latency in Table~\ref{tab:flowsr_efficiency}.
The MACs and runtime are measured for 4$\times$ SR using a 128$\times$128 LR input.
Note that we use a fixed text prompt for model inference, eliminating the need for text encoding in the SD model.
As demonstrated, our method shows a significant advantage over multi-step SR approaches, such as StableSR and SeeSR, while maintaining comparable computational complexity to one-step methods like OSEDiff.

\begin{table}[htbp]
\caption
{Efficiency metrics of parameters, MACs, and runtime.}
\label{tab:flowsr_efficiency}
\begin{adjustbox}{width=\hsize}
\begin{tabular}{cccccccccc}
\toprule
Method                   & StableSR & DiffBIR & SeeSR & PASD  & ResShift & SinSR & OSEDiff & DoSSR & FlowSR \\ \midrule
\#steps $\downarrow$     & 200      & 50      & 50    & 20    & 15       & 1     & 1       & 1     & 1      \\
\#param (M) $\downarrow$ & 1409     & 1683    & 2511  & 2314  & 174      & 174   & 1765    & 1718  & 982    \\
MACs (G) $\downarrow$    & 95382    & 24234   & 66444 & 23592 & 4962     & 2119  & 2323    & 3232  & 2148   \\
time (s) $\downarrow$    & 13.54    & 6.51    & 5.21  & 3.47  & 0.89     & 0.13  & 0.16    & 0.28  & 0.14   \\ \bottomrule
\end{tabular}
\end{adjustbox}
\end{table}

\subsection{More Qualitative Visual Comparisons}

\begin{figure*}[t]
    \centering
    \includegraphics[width=0.96\hsize]{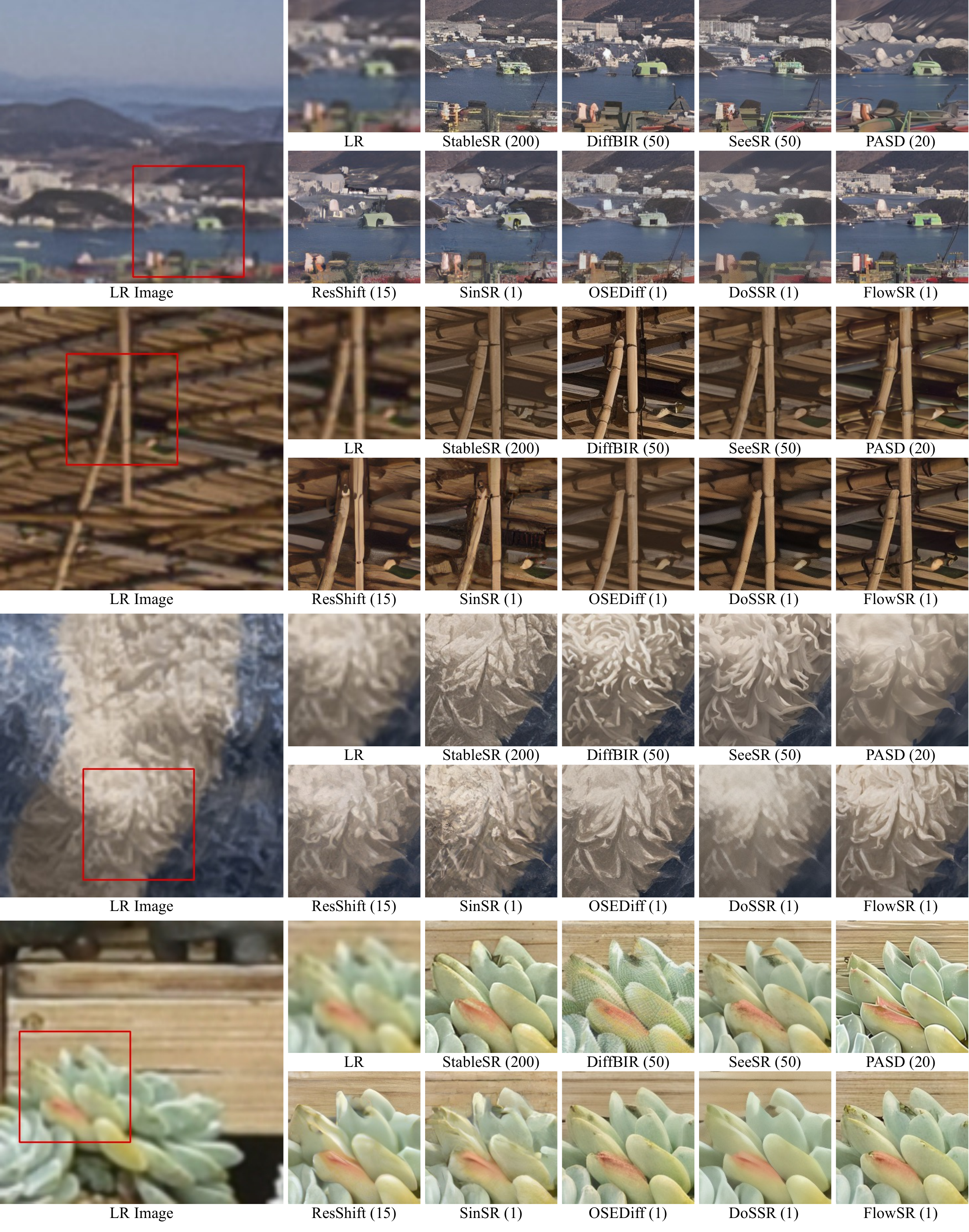}
    \caption
    {Visual comparisons of different SR methods on real-world examples \#1. 
    The number of sampling steps are indicated in bracket. Please zoom in for a better view.}
    \label{fig:flowsr_vis_supp_1}
\end{figure*}

\begin{figure*}[t]
    \centering
    \includegraphics[width=0.96\hsize]{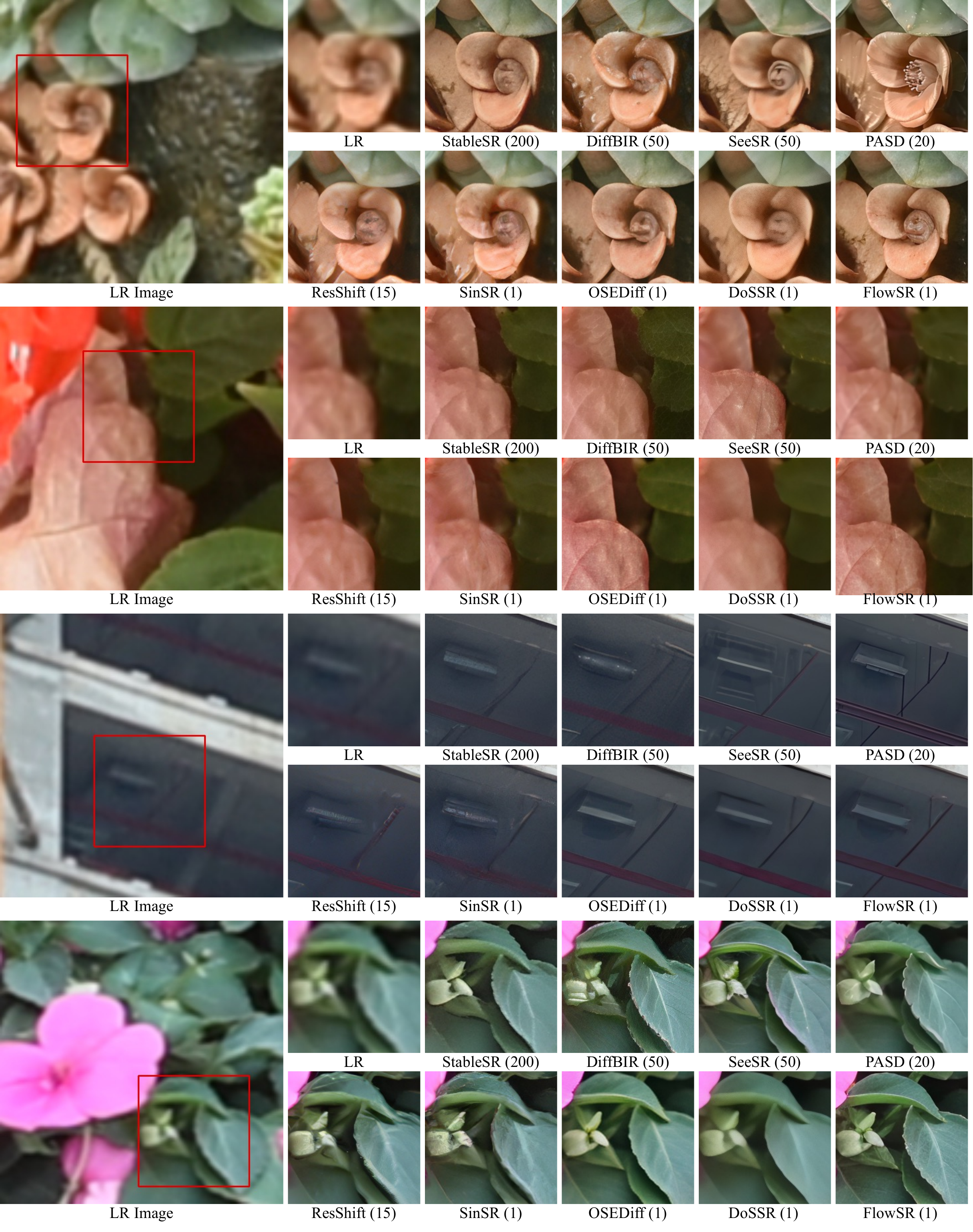}
    \caption
    {Visual comparisons of different SR methods on real-world examples \#2. 
    The number of sampling steps are indicated in bracket. Please zoom in for a better view.}
    \label{fig:flowsr_vis_supp_2}
\end{figure*}

\begin{figure*}[t]
    \centering
    \includegraphics[width=0.96\hsize]{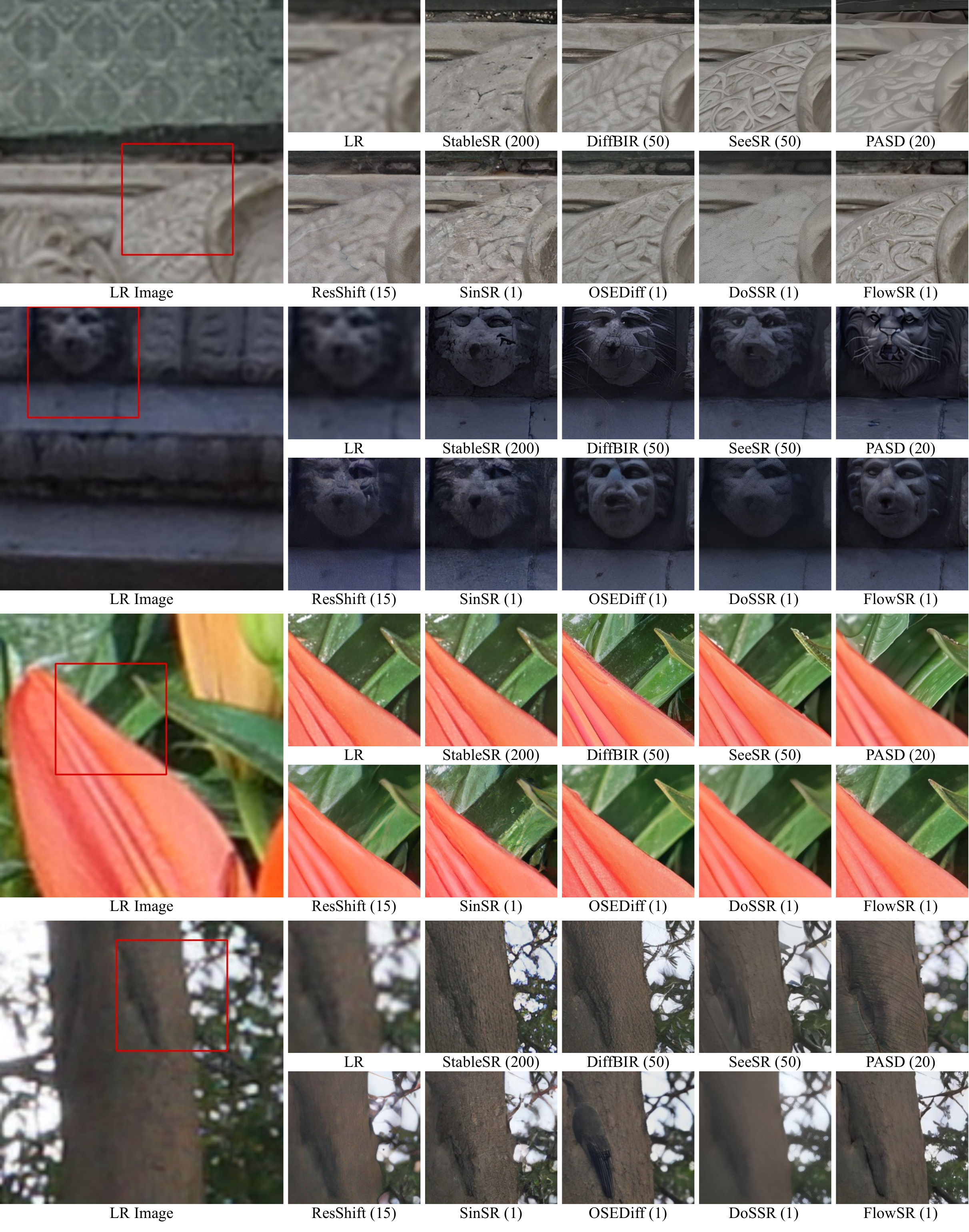}
    \caption
    {Visual comparisons of different SR methods on real-world examples \#3. 
    The number of sampling steps are indicated in bracket. Please zoom in for a better view.}
    \label{fig:flowsr_vis_supp_3}
\end{figure*}

\cref{fig:flowsr_vis_supp_1,fig:flowsr_vis_supp_2,fig:flowsr_vis_supp_3} provide additional visual comparisons between FlowSR and other DM-based SR methods.
Our visual results are consistently better than, or at least comparable to, all multi-step and single-step diffusion methods across various scenarios, such as flowers, buildings, and clothing.
Visual comparisons also support the conclusions drawn from the quantitative study, highlighting the higher fidelity of our results.
Overall, FlowSR exhibits more natural details, along with realistic textures and structures.

\begin{figure}[t]
    \centering
    \includegraphics[width=\hsize]{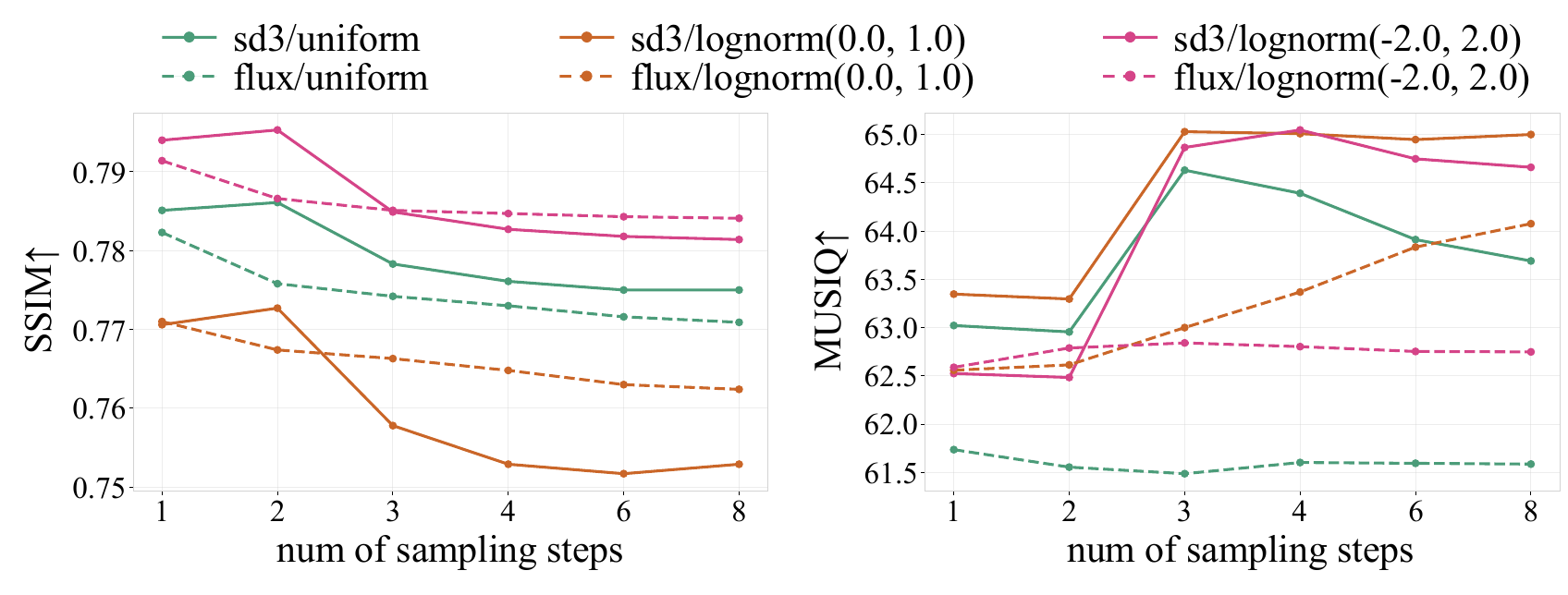}
    \caption
    {Impact of timestep shifting / timestep sampling.
    SD3 timestep shifting with \texttt{lognorm(-2.0, 2.0)} timestep sampling achieves a good fidelity/quality tradeoff on DRealSR \cite{wei2020component}.}
    \label{fig:ablation_time}
\end{figure}

\subsection{Impact of timestep shifting and sampling}
\label{subsec:flowsr_supp_timestep}

We train the basic SR flow models using different time scheduling methods to evaluate their impact.
We select representative timestep shifting options, including SD3 \cite{esser2024scaling}, which biases timesteps toward $t=1$, and FLUX.1-schnell\footnote{\url{https://huggingface.co/black-forest-labs/FLUX.1-schnell}}, which uses uniform timesteps.
For timestep sampling, we use \texttt{lognorm(0.0, 1.0)} as adopted in \cite{esser2024scaling}, \texttt{lognorm(-2.0, 2.0)} studied in \cite{sauer2024fast}, and uniform sampling.
The first sampling method favors intermediate timesteps, while the second samples more timesteps closer to $t=1$.
The results for different inference steps are shown in \cref{fig:ablation_time}.
We observe that: (1) SD3 timesteps outperform the uniform timesteps for SR flow in most cases; (2) \texttt{lognorm(0.0, 1.0)} achieves high quality (MUSIQ) but sacrifices fidelity (SSIM).
In our experiments, we employ SD3 timesteps with \texttt{lognorm(-2.0, 2.0)} timestep sampling, as it demonstrates high fidelity with one-step inference and good quality with few-step inference.

\section{Limitations and Future Works}
\label{sec:flowsr_supp_limitation}

In this work, we tackle one-step SR from the perspective of flow and consistency.
We provide valuable insights into the effective use of flow-based techniques and consistency learning to achieve competitive SR results in a single-step setting.
While our study demonstrates promising results, there are some limitations.
First, due to computational constraints, we have not yet explored more advanced T2I models, such as SD3 \cite{esser2024scaling} and FLUX \cite{flux2024}, as potential backbones.
Second, we are actively working on further reducing the number of parameters in the backbone network to achieve additional efficiency gains.

\end{document}